\documentclass[10pt,letterpaper]{IEEEtran}
\PassOptionsToPackage{bookmarks={false}}{hyperref}
\usepackage{amssymb, amsmath}
\usepackage{multirow}
\usepackage{rotating}
\pdfoutput=1

\ifCLASSINFOpdf
\else
\fi
\begin{document}

\title{High Frequency Content based Stimulus for Perceptual Sharpness Assessment in Natural Images}

\author{Ashirbani~Saha,~\IEEEmembership{Student Member,~IEEE,}
        and Q.~M.~Jonathan~Wu,~\IEEEmembership{Senior Member,~IEEE}
        }


\maketitle

\begin{abstract}
A blind approach to evaluate the perceptual sharpness present in a natural image is proposed. Though the existing literature demonstrates a set of variegated visual cues to detect or evaluate the absence or presence of sharpness, we emphasize in the current work that high frequency content and local standard deviation can form strong features to compute perceived sharpness in any natural image, and can be considered an able alternative for the existing cues. Unsharp areas in a natural image happen to exhibit uniform intensity or lack of sharp changes between regions. Sharp region transitions in an image are caused by the presence of spatial high frequency content. Therefore, in the proposed approach, we hypothesize that using the high frequency content as the principal stimulus, the perceived sharpness can be quantified in an image. When an image is convolved with a high pass filter, higher values at any pixel location signify the presence of high frequency content at those locations. Considering these values as the stimulus, the exponent of the stimulus is weighted by local standard deviation to impart the contribution of the local contrast within the formation of the sharpness map. The sharpness map highlights the relatively sharper regions in the image and is used to calculate the perceived sharpness score of the image. The advantages of the proposed method lie in its use of simple visual cues of high frequency content and local contrast to arrive at the perceptual score, and requiring no training with the images. The promise of the proposed method is demonstrated by its ability to compute perceived sharpness for within image and across image sharpness changes and for blind evaluation of perceptual degradation resulting due to presence of blur. The experiments conducted on four publicly available databases demonstrate improved performance of the proposed method over that of the state-of-the-art techniques.
\end{abstract}

\begin{IEEEkeywords}
High pass filter, Perceptual sharpness, local standard deviation
\end{IEEEkeywords}

\section{Introduction}
\label{sec:Introduction}
The sharpness present in an image is indicative of the manifestation of the fine details present in it. When the term perceived sharpness is used, we essentially refer to the amount of sharpness that is perceived by an human observer from the image. Sharpness is reduced in an image by the presence of blur, which can be naturally introduced by motion or varying focus and also artificially introduced by filtering techniques. Also, sharpness can be increased in any image naturally by focussing properly or artificially by unsharp masking, image deblurring etc. Perceived sharpness assessment can be considered as a special type of image quality assessment (IQA) that considers the degradation of pristine quality image mainly due to the presence of blur~\cite{FerzliBlur2009}. IQA has been gaining a lot of interest since the last decade due to its applicability in multimedia systems. Though, the quality of an image can be readily assessed by the human eyes, it is indeed difficult to implement an automated process to do so. Different types of distortions and their varying effects on the human visual system (HVS) are the main reasons behind such difficulty. Availability of a pristine quality image makes this job easier~\cite{WangIQABook} and the IQA technique that needs the full information of a pristine quality image to do so is called a full-reference (FR) measure. When no information is used by an IQA technique about the pristine quality image, the technique is called No-reference (NR). Though humans can assess the quality of an image without any reference~\cite{WangIQABook}, numerous types of distortions, partial information about the working of HVS and above all absence of the reference image are the principal challenges towards developing a robust automatic technique. However, if the distortion and its nature become known, specific visual cues and information can be used to describe the image. The NR-IQA techniques which can evaluate perceptual degradations caused by a single distortion can be denoted by the term NR-Specific. NR-General methods are those which work on a variety of distortions. In this work, we concentrate on the perceptual degradation caused by lack of image sharpness or due to the presence of blur. Quantifying the perceptual degradation caused by lack of sharpness is an important task and becomes particularly relevant in evaluating several image processing applications. The perceptual sharpness measure represents any image with a scalar value signifying the amount of sharpness present in it. Thus, image enhancement techniques~\cite{Shaked2005} that require performance evaluation related to sharpness can also use these measures. Image compression, deblurring, restoration, registration and different applications requiring quality evaluation may not have the reference image available. Hence, NR or blind evaluation of sharpness becomes necessary.

Over the years, several techniques related to the detection and analysis of blur or sharpness have been proposed. However, objective techniques related to the assessment of perceptual degradation caused by lack of sharpness are relatively newer. Grammatically, the antonym of blur is sharpness and several blur or sharpness detection measures are available in the existing literature. However, as explained in~\cite{VuS32012}, the absence of blur does not imply presence of sharpness from the image processing point of view. A uniform region is neither sharp nor blur but high frequency content is not present. Presence of sharpness implies existence of high frequency content~\cite{GonzalezWoods2002}. It is widely agreed that the attenuation of high frequency content can cause an image to get blurred~\cite{VuS32012}. Thus, in the approach of Vu et al.~\cite{VuS32012}, spectral slope was used to measure the attenuation of the high frequency content. In addition to this, total variation was used as a measure of contrast to define the perceptual quality. The approach used both spectral and spatial domains to generate the quality score, thus proving the importance of high frequency content and contrast measure. The contrast measure can also be thought of as a measure of the local content of the given image. As seen in~\cite{FerzliBlur2009}, the local content of image is also responsible for the perceived degradation caused by blur.

Motivated by the aforementioned works, we hypothesize that high frequency content, in form of high pass filtered image, can be used as the principal stimulus for evaluation of degradation caused by blur. The contrast information in the form of local standard deviation can be used as weights for the high frequency content. Primarily, simple high pass filtering, which readily shows the pixels responsible for the sharpness or high frequency content present in the image, is used in the proposed technique. The local standard deviation is used to weigh the filtered image. Since, high pass filtered image is the principal stimulus, an exponent of the filtered and weighted image pixels is practically used for the proposed measure. Finally, the sharpness map is constructed with the help of logarithm of the filtered and weighted image. We have carried out extensive experiments to demonstrate the dependence of the performance on the exponent. We have further evaluated the performance of the proposed approach using qualitative and quantitative analysis to demonstrate the promise it delivers. A comparative study with the existing techniques is carried out using four publicly available databases. We have also used the high frequency content obtained from the high pass Gaussian filter and undecimated wavelet transform to demonstrate the ability of high frequency content obtained from both of these techniques to serve as the stimulus.

The rest of the paper is organized as follows. In Section~\ref{sec:PreviousWorks}, we discuss about the existing approaches related to the evaluation of blur. The proposed method is discussed in details in Section~\ref{sec:ProposedMethod}. The experiments related to the proposed technique are presented in Section~\ref{sec:ExperimentsResults}. Finally, we conclude in Section~\ref{sec:Conclusion}.

\section{Overview of Existing Techniques}
\label{sec:PreviousWorks}
The existing techniques in the literature involve several visual cues to detect and evaluate sharpness or blur based degradation without the aid of any reference image. The set of variety of visual cues used for sharpness/blur detection is quite large. Depending on the type of features used to evaluate sharpness/blur, the techniques can be classified as spatial domain, transform domain and hybrid techniques. Though all of the approaches mentioned below do not fall exactly into the category of quantifying perceptual distortions formed due to lack of sharpness/presence of blur, they provide information about different types of visual cues that aid in detecting blur or sharpness and the variety of techniques that are developed over several decades to evaluate blurriness or sharpness.
\subsection{Spatial domain based techniques}
\label{ssec:PreviousWorksSpatial}
Spatial domain sharpness/blur detection methods rely solely on features extracted or pixel properties from the image in the spatial domain only. Image variance from entire image is used as a measure in~\cite{ErasmosVariance1982} as variance is likely to decrease in an image with the presence of blur. A globally sharp image has lower correlation between adjacent pixels. Thus, when the previous pixel is predicted from the current one, the prediction error is likely to be higher. The variance of this error is used to determine a globally blurred or sharpened image~\cite{KimBlur2008}. Since, neighboring pixels are more correlated in a blurred image compared to its sharper version, an autocorrelation function based measure is proposed in~\cite{Batten2000}. Image intensity histograms~\cite{JarvisBlur1976,Krotkov1989} based measures are also used to determine blur in auto-focussing applications. By discriminating between different levels of blur on the same image, a no-reference blur measure is proposed by Crete~\cite{CreteBlur2007}. However, the image edge or gradient based features happen to be widely used for determination of perceptual blur. The edge feature based methods only rely on the edge width or the spread of edges. The work leading to this approach is~\cite{Marziliano2002}where the width of an edge (vertical or horizontal) is determined by separation of the extrema positions closest to any edge. This width is considered to be the local edge value and the average of such values over an image serves to be the measure of the blurriness of the image. The width of edge determined using the local gradient direction is used to measure blurriness in~\cite{Ong2003}. Degradations caused by the blur uniformly over an whole image is also evaluated using Renyi entropy based anisotropy measure~\cite{GabardaAnisotropy2011}. However, depending on the image content, the perceived blurriness of an image varies and this was incorporated through the concept of just noticeable blur (JNB) in the edge width~\cite{FerzliBlur2009} to determine the probability of blur detection for each block region in an image. Finally, a Minkowski metric based pooling is used to determine overall perceived blur and the measure is named as just noticeable blur measure (JNBM). A modification to the same approach using cumulative probability of blur detection (CPBD) is proposed in~\cite{NarvekarCPBD2011}. Combination of edge based features and variance are used to obtain a sharpness measure in~\cite{Chung2004}. Based on the argument that singular value decomposition of local gradient values represent the components along the direction of significant change of gradient direction, a sharpness based measure called H-metric was proposed in~\cite{ZhuBlur2009}.
\subsection{Transform domain based techniques}
\label{ssec:PreviousWorksTransform}
Transform domain approaches began with the knowledge of failure of blurred image spectra to follow the power law as done by natural images~\cite{FieldBrady1997}. The kurtosis obtained from the Fourier Spectra was used to measure the image sharpness in~\cite{Zhangsharp2003}. Some blur measures using high and low frequency power measures are presented in~\cite{Firestone1991}. A DCT based approach was proposed in~\cite{Marichalblur1999}. 2D DCT coefficient based kurtosis is used to calculate the image sharpness in~\cite{Caviedes2004}. Wavelet transform based blur detection approaches are presented in~\cite{Tongblur2004, VuSharpness2012}. The techniques discussed so far did not involve any kind of prior training.  A wavelet feature based SVM classification and related confidence was used to design the quality score of degraded images in~\cite{ChenBlur2011}. A complex wavelet based computation of local phase coherence~\cite{HassenSharpness2013}, called LPC-SI, is shown to perform well in four databases among the recent approaches developed for perceptual sharpness detection.
\subsection{Hybrid techniques}
\label{ssec:PreviousWorksHybrid}
Hybrid techniques rely on both spatial domain and transform domain methods to evaluate blur or sharpness. The work proposed by Sadaka et. al~\cite{SadakaBlur2008} used JNB and incorporated foveal pooling into it by using Gabor filtering based visual attention map. A combination of features obtained from local power spectrum, gradient histogram span and color based maximum saturation was used to detect blur regions in a partially blurred image and then used to classify these blur regions into motion and out-of-focus blurred regions in~\cite{RentingCVPR2008}. A recent approach, named S3~\cite{VuS32012}, uses both spatial and spectral information to perceptually evaluate blur. In addition, they also come up with a blur map to identify the sharp and blurred regions in an image. Apart from these techniques, the general purpose image quality assessment techniques like BRISQUE~\cite{MittalBRISQUE2012} and BLIINDS II~\cite{SaadBLIINDS2011} also evaluate degradation caused by the presence of blur. The proposed technique can also be considered hybrid as the properties from both of the transform domain and spatial domain are used to design it, though the calculations are carried out spatial domain only.
\section{Proposed Method}
\label{sec:ProposedMethod}
\begin{figure*}
\begin{center}
  \includegraphics[width=0.9\textwidth]{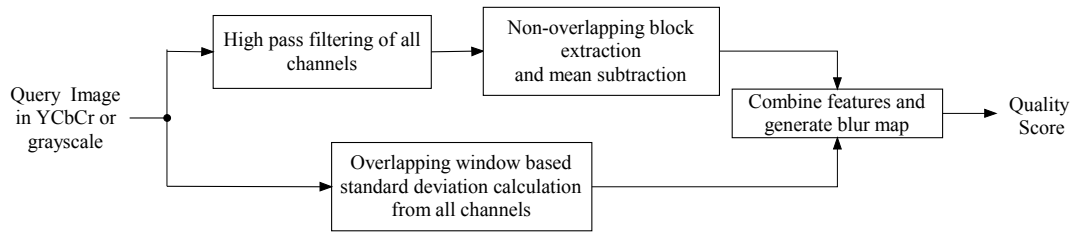}\\
  \caption{Block Diagram of the proposed method}\label{Fig:ProcessDiagram}
  \end{center}
\end{figure*}
In this section, the proposed method is discussed in details. The block diagram of the method is presented in Fig.~\ref{Fig:ParameterAndDatabases}. We describe the proposed method for a color image for generalization. In case of a grayscale image, the steps for the color image are performed for single channel only. Let $I_c$ be the color image whose perceptual sharpness is to be determined. The pixel values of the image are scaled down between 0 and 1. At first, the image is transformed to YCbCr color space from $\textrm{RGB}$  with the channels $I_Y$ denoting luminance and $I_{Cb}$ and $I_{Cr}$ denoting the chrominance channels. This color space transformation is carried out as YCbCr space exhibits improved perceptual uniformity with respect to $\textrm{RGB}$~\cite{ImageDatabase2002}.  Next, we have used either high pass filter (HPF) derived by subtracting Gaussian low pass filter (of width 3$\times$3 and standard deviation 0.25) from all pass filter or the undecimated wavelet transform (UWT) for extracting the high frequency content. These two filters are chosen as they form very basic and simple techniques for extracting high frequency content. For the HPF, the filter is denoted by $H_{G}$ in the spatial domain and standard deviation of the low pass filter from which it is derived is indicative of its bandwidth. As calculated, the cut off frequency of 1D high pass filter derived from the Gaussian low pass filter of standard deviation 0.25 is $\frac{\pi}{4.934}$ rad/s. For HPF, the high frequency content is then obtained by
\begin{align}
H\_I_Y &= H_G \otimes I_Y,\\
H\_I_{Cb} &= H_G \otimes I_{Cb},\\
H\_I_{Cr} &= H_G \otimes I_{Cr},
\end{align}
where $\otimes$ denotes the two-dimensional convolution operation between two signals. When UWT is used for extracting high frequency content, the diagonal subbands from the single level UWT decomposition of all of the channels are chosen as $H\_I_Y$, $H\_I_{Cb}$ and $H\_I_{Cr}$. The diagonal subbands are chosen as they are result of processing by two high pass filters whereas each of other subbands are formed due to the application of at least one low pass filter~\cite{StarkUWT2007}. The MATLAB implementation of two-dimensional UWT using `db1' has been used for our purpose. Next, from each of these matrices locating and indicating high frequency content, the non-overlapping blockwise means of the high frequency content is subtracted and the absolute values of the differences are used to form matrices $MH\_I_Y$, $MH\_I_{Cb}$ and $MH\_I_{Cr}$ from $H\_I_Y$, $H\_I_{Cb}$ and $H\_I_{Cr}$ respectively. This process removes the blockwise high frequency content bias. The block size $b_l$ used is fixed in all of our experiments.
Now the local standard deviation with the same block size is calculated for all overlapping blocks from the image channels to obtain the matrices $S\_I_Y$, $S\_I_{Cb}$ and $S\_I_{Cr}$ from $I_Y$, $I_{Cb}$ and $I_{Cr}$ respectively. The local standard deviation serves as the contrast measure in natural images~\cite{Bex02}. The standard deviation is used to weigh the high frequency content to obtain the matrices $T_Y$, $T_{Cb}$ and $T_{Cr}$. Each element ($i$,$j$) of the matrices is calculated as follows:
\begin{align}
T_Y(i,j) &= \frac{[ MH\_I_Y(i,j)]^\alpha \times S\_I_Y(i,j)}{\sum_{i}\sum_{j}(S\_I_Y(i,j))}\\
T_{Cb}(i,j) &=  \frac{[ MH\_I_{Cb}(i,j)]^\alpha \times S\_I_{Cb}(i,j)}{\sum_{i}\sum_{j}(S\_I_{Cb}(i,j))}\\
T_{Cr}(i,j)&=  \frac{[ MH\_I_{Cr}(i,j)]^\alpha \times S\_I_{Cr}(i,j)}{\sum_{i}\sum_{j}(S\_I_{Cr}(i,j))}
\end{align}
Here, $\alpha$ denotes the exponent for high frequency content obtained from each channel.
We know from the aforementioned equations that, if the local contrast is high enough, the high frequency content is weighted more. Now the total stimulus due to the high frequency content weighted by local standard deviation is calculated as \begin{equation}
\label{Eqn:Tot_Stm}
TS(i,j) = \left(\frac{T_Y(i,j) + T_{Cb}(i,j) + T_{Cr}(i,j)}{3}\right)^\frac{1}{\alpha}
\end{equation}
Now the raw sharpness map is generated by the following relation:
\begin{equation}
\label{Eqn:Raw_map}
S_{map}(i,j) = \frac{\textrm{abs}\left[\textrm{log}(\epsilon)  + \epsilon \right]}{\textrm{abs}\left[\textrm{log}(TS(i,j)) + \epsilon)  + \epsilon \right]},
\end{equation}
where $\epsilon$ is a small positive number used to avoid instability during taking logarithm and performing division.
The values $T_Y(i,j),~T_{Cb}(i,j)$ and $T_{Cb}(i,j)$ are always less than 1 and greater than zero. Hence natural logarithm of their sum is negative. The closer the sum is to zero, the more is the absolute value of the logarithm. For sharper regions, higher values for contrast and high pass filtered content are expected. Hence, the sum is closer to 1 and $S_{map}(i,j)$ is high. For blurred images, the sum will be closer to zero and hence, the denominator will be high leading to lower values of $S_{map}(i,j)$.
In order to generate the sharpness score, the border of the sharpness map is discarded depending on the block size, hence eliminating the border effect; the remaining map is called $BS_{map}$. The perceptual quality score is finally obtained as
\begin{equation}
Q_s = \left(\textrm{max}(BS_{map})\right).
\end{equation}
where max(.) calculates the maximum value of the argument. The max(.) operator here closely follows the approach of the HVS to combine visual responses in a non-linear fashion~\cite{Graham1989} and thus we apply maximum pooling to arrive at the sharpness score. The $BS_{map}$ lacks localization and hence we calculate the final localized map by median filtering. In order to enhance the visualization of the sharper regions along with localization, we do the following. First, we calculate $\gamma$ as
\begin{equation}
\gamma = \frac{\textrm{max}(BS_{map}) + \epsilon}{\textrm{mean}(BS_{map}) + \epsilon}.
\end{equation}
The final localized blur map is calculated by
\begin{equation}
LBS_{map} = \exp(\gamma\times\textrm{block\_median}(BS_{map},b_l + 2))
\end{equation}
where block\_median($A,n$) calculates overlapping blockwise median of the matrix $A$ for a block size given by $n \times n$. Now we present two discussions on the intuitive explanation and the parameters of the proposed method.
\subsubsection{Intuitive explanation of the proposed method}
\label{sec:IntuitiveExplanation}
The proposed method actually echoes the psychometric function presented in~\cite{Robson1981} which expresses the probability $P_t$ of detection of a stimulus $S_t$ by detector $t$ is dependent on the contrast $C_t$ related to the stimulus as
\begin{equation}
\label{Eqn:Prob}
P_t = 1 - 2^{(-S_tC_t)^q},
\end{equation}
where $q$ is the exponent.
From this equation, it is revealed that the product $(S_tC_t)^q$ is of major importance in calculating the probability of detecting the stimulus. The stimulus $S_t$ in our proposed method is the high frequency content derived after the high pass filtering of the signal. We do a basic thresholding in the stimulus by subtracting the blockwise mean as discussed earlier in this section. The $C_t$ is obtained in the form of the local standard deviation. For each channel we calculate the product of the exponentiated high frequency content and local standard deviation to obtain $T_Y$, $T_{Cb}$ and $T_{Cr}$. The exponent $\alpha$ can be compared to the exponent $q$ in Eqn.~\ref{Eqn:Prob}. However, it differs from $q$ is one aspect; $\alpha$ values are used to exponentiate the principal stimulus only. After that, a summation of these values takes place to generate a score. The final form of the measure in Eqn.~\ref{Eqn:Raw_map}, is used to increase the range of the values obtained from proposed method. The most important part is therefore the argument of the logarithm present in the denominator for calculating raw sharpness map in this equation. As we designed the measure, the maximum value of the sharpness map $BS_{map}$ is selected as representative perceived sharpness. Hence, it is theoretically ensured in the final form that the value of the measure increases with the perceived sharpness.
\subsubsection{Discussion about the parameters involved}
\label{sec:ProposedMethodParameter}
\begin{figure*}
\begin{center}
  \includegraphics[width=0.9\textwidth]{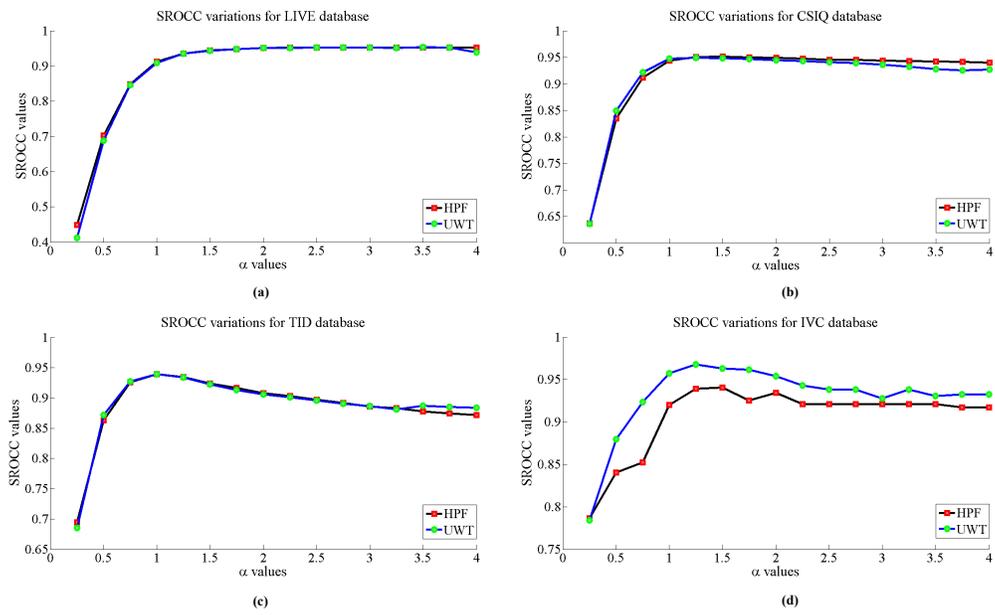}\\
  \caption{Variation of SROCC scores for different values of $\alpha$ using HPF and UWT}\label{Fig:ParameterAndDatabases}\end{center}
\end{figure*}

The block size $b_l$ and map exponent $\alpha$ two main parameters of the implementation. The block size $b_l$ is used in several computations and a fixed value of 7 is used for it in every experiment. Larger values of $b_l$ increases the computation time. $\alpha$ is the parameters which control the contribution of stimulus to determine the final quality score. Hence, we have carried out experiments with the $\alpha$ values. The effect of change of these values are shown in Fig.~\ref{Fig:ParameterAndDatabases}. As we see, the performance improves remarkably for $\alpha$ value greater than 1. Since, the elements of the matrices $MH\_I_Y$, $MH\_I_{Cb}$ and $MH\_I_{Cr}$ lie between 0 and 1, an exponent value less than 1 increases the values. Exponent values greater than decreases their values. We find from the experiments that better performance are achieved if the values of the stimulus remain same or decrease. In the experiments, $\alpha=2$ is used as for $\alpha$ values greater than 1, high and less fluctuating SROCC values are obtained for all databases.

\section{Experiments and Results}
\label{sec:ExperimentsResults}
In this section, we discuss several experiments conducted using the proposed method. First, we demonstrate the qualitative and quantitative results on perceived sharpness for different sharpness levels on the same image. Secondly, we discuss the performance of our measure on a set of different images with varying amount of blur. In the third experiment, the performance in four databases are presented. A comparative study with several existing methods are presented in the fourth experiment. Four databases are used for the experiments: LIVE database~\cite{SheikDB}, CSIQ database~\cite{Larson10}, TID2008 database~\cite{PonomarenkoTID2008} and IVC database~\cite{ivcdb}. Pristine and corresponding blurred images are available in these databases along with the subjective scores provided by the human observers. The subjective scores are given in the form of mean opinion scores (MOS) or differential mean opinion scores (DMOS). The following evaluation measures are used for analysis and comparison of the performance of the proposed measure: Spearman's Rank-Order Correlation Coefficient (SROCC), Kendall's Rank-Order Correlation Coefficient (KROCC), Pearson's Linear Correlation Coefficient (PLCC) and Root Mean Square Error (RMSE). Among these, SROCC is the most used measure and it calculates the prediction monotonicity of a perceptual quality measure. KROCC is also another evaluator of monotonicity. PLCC measures prediction accuracy. Each of SROCC, KROCC and PLCC can have maximum value as 1 and the higher they are, the better is the quality of the distortion measure. RMSE is a measure of error with lower values indicating better quality of performance of the measure, while its maximum value is dependent on the range of the subjective quality scores in the database. PLCC and RMSE are calculated after mapping the objective scores obtained from the image quality measures with MOS/DMOS values provided in the databases using a 5-parameter logistic function~\cite{Sheik06}.
\subsection{Different blur levels in same pristine image}
\label{ssec:sameImageBlurNRF}
\begin{figure*}
\begin{center}
  \includegraphics[width=0.9\textwidth]{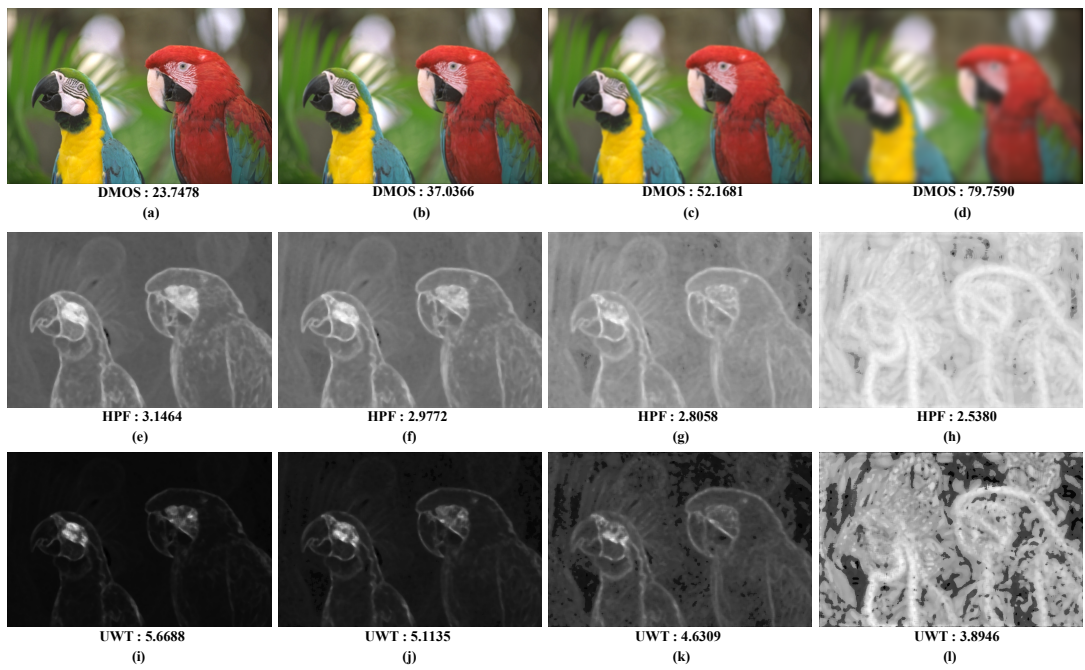}\\
  \caption{Sharpness maps and quality score for different levels of blur over the same pristine image from LIVE database for HPF and UWT. Each column in the figure represents the query image, and the blur maps resulting from HPF and UWT respectively considering the direction from top to bottom.}\label{Fig:DifferentBlurLevels}\end{center}
\end{figure*}

In the first experiment, we select a pristine image from the LIVE database and consider its several blurred versions. As per~\cite{SheikDB}, varying levels of blur are applied on the pristine image `Parrot' to generate four different images which are shown in the first row of Fig.~\ref{Fig:DifferentBlurLevels}. The figures in the first row are arranged in the increasing order of the corresponding DMOS values, signifying the degradation of perceptual quality resulting due to blur or loss of sharpness. In the first two images, the birds can clearly be identified as the part of the foreground. However, the distinction between the foreground and background gradually diminishes as the DMOS increases. The next two rows in the same figure shows the corresponding sharpness maps obtained using HPF and UWT respectively. A closer look in each of these images show that the relative sharpness values at each pixel are shown in the sharpness maps. The range of the values in the maps are adjusted for better visualization. The maps produced by HPF and UWT are different as sharper regions are more discriminated by UWT when compared to HPF. In Fig.~\ref{Fig:DifferentBlurLevels}(a), the birds' outlines and eye regions are very sharp compared to the background. As we proceed over from Fig.~\ref{Fig:DifferentBlurLevels}(b) to Fig.~\ref{Fig:DifferentBlurLevels}(e), the sharpness difference between the backgrounds and the birds drops gradually signifying the loss of relative sharpness between the background and foreground. The score of the proposed measure goes down in the same order as that of the DMOS values. This experiment shows that the proposed measure follows the rank order of the DMOS scores for different levels of blur applied on the same image. This experiment also shows that for varying levels of blur applied on the same image, the maps produced are able to highlight relatively sharper regions.
\subsection{Different blur levels across different pristine images}
\label{ssec:diffImageBlurNRF}
\begin{figure*}
\begin{center}
  \includegraphics[width=0.9\textwidth]{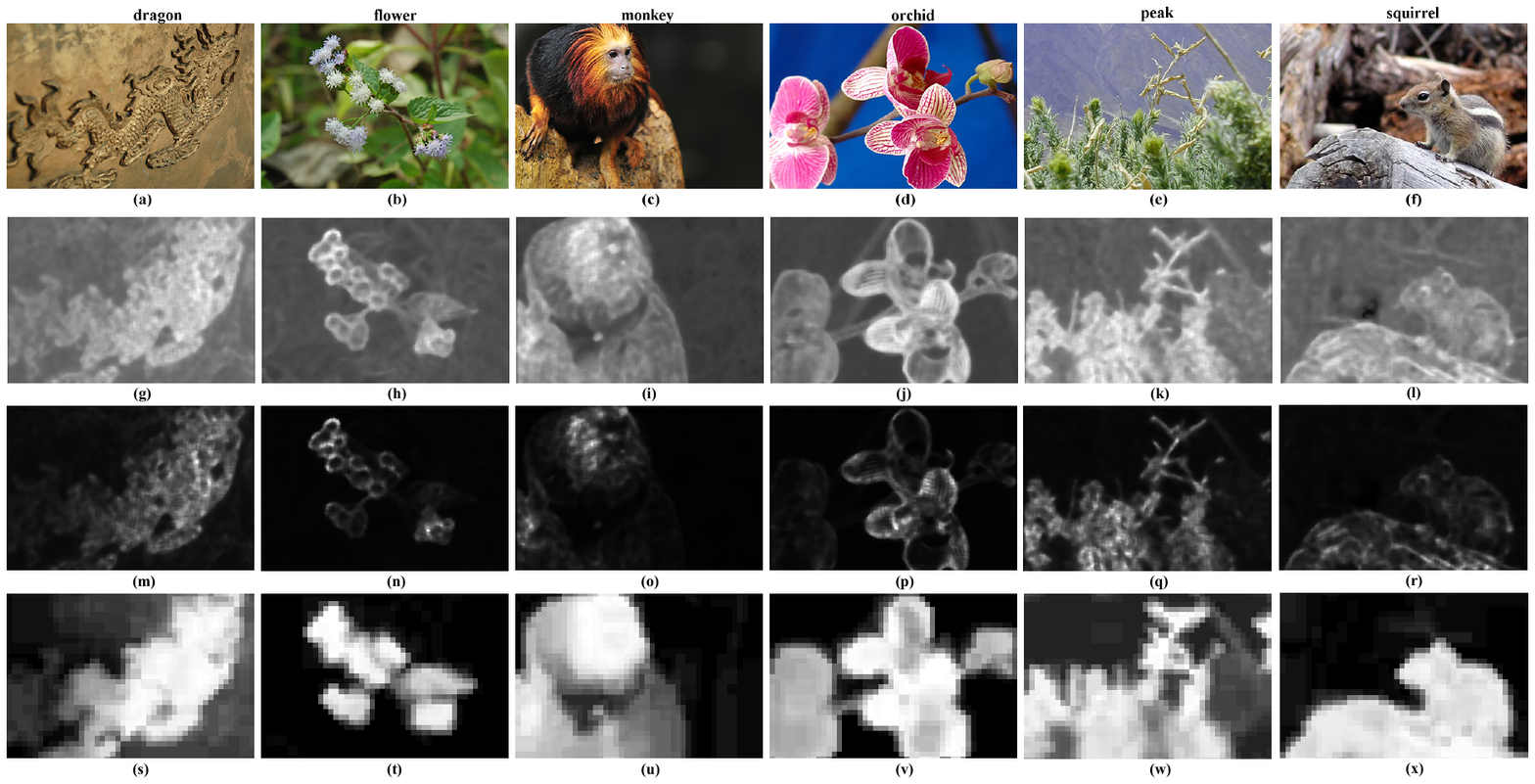}\\
  \caption{Sharpness maps for different images in~\cite{S3maps}. Each of the six columns contains the query image. the sharpness map obtained from HPF and UWT respectively and finally the ground truth from top to bottom.}\label{Fig:BlurMapsS3DatasetWOSCORES}
\end{center}
\end{figure*}

The second experiment involves the performance evaluation of the proposed measure using different pristine images. The first part of the experiment is qualitative and second part is quantitative in nature. In the first part of the experiment, we test the performance on the suite of images provided with the implementation of S3~\cite{S3maps}. These images have different areas with different levels of sharpness. The images and corresponding sharpness maps are presented in Fig.~\ref{Fig:BlurMapsS3DatasetWOSCORES}. The first row of the figure presents the original images with the following rows representing the maps obtained from the proposed measure using HPF and UWT and the ground truth respectively. However, one point needs to be clarified, while comparing the maps obtained from proposed method to the ground truth. The ground truths were generated using the evaluation of the subjects on 16$\times$16 non-overlapping blocks of the image whereas the maps obtained from the proposed method are more localized. Therefore, a comparison can be made in the following way: the right side of the Fig.~\ref{Fig:BlurMapsS3DatasetWOSCORES}(a) `dragon' had relatively higher sharpness than that of the left side. The map from the proposed method supports that. As we can see the proposed methods using HPF and UWT are able to identify the relative sharpness and blur in the images. In the Fig.~\ref{Fig:BlurMapsS3DatasetWOSCORES}(b), `flower', the flowers in the middle have considerably higher sharpness compared to the background. The map in Fig.~\ref{Fig:BlurMapsS3DatasetWOSCORES}(h) supports it but as it is localized, it is able to distinguish between the sharpness levels at the  borders and middle portion of the flower. Fig.~\ref{Fig:BlurMapsS3DatasetWOSCORES}(n), obtained using UWT also reflects the sharper areas in the flower image. In the `monkey' image from Fig.~\ref{Fig:BlurMapsS3DatasetWOSCORES}(c), the left paws, fur closest to the monkey's face, monkey's eyes and fur close to the right leg are sharper areas. The same is supported by the map obtained using HPF in accordance with the ground truth. For the same image using UWT, the fur, eyes and paws are considered much sharper than the rest of the images. As seen from the ground truth of the `orchid' image shown in Fig.~\ref{Fig:BlurMapsS3DatasetWOSCORES}(d), the petals in the middle have higher sharpness than that of the petals on the left. This is also captured by the maps obtained from the proposed approach. The `peak' image as shown in Fig.~\ref{Fig:BlurMapsS3DatasetWOSCORES}(e) has a blurred background, and the plants at the middle and left have more sharpness than the background. From the map generated by the proposed methods, we can observe the same. Moreover, the plants at right have sharpness more than the background and less than the plants in middle. This is also supported by the ground truth. Finally, the `squirrel' image in Fig.~\ref{Fig:BlurMapsS3DatasetWOSCORES}(f) shows the wooden log and the squirrel has more sharpness compared to the background. The eye, borders of the ears and paws are sharper as shown by the proposed measures and the ground truth. This experiment makes clear that the maps generated using UWT demonstrate more contrast between the sharp and unsharp parts as compared to the maps generated using HPF.

\begin{figure*}
\begin{center}
  \includegraphics[width=0.9\textwidth]{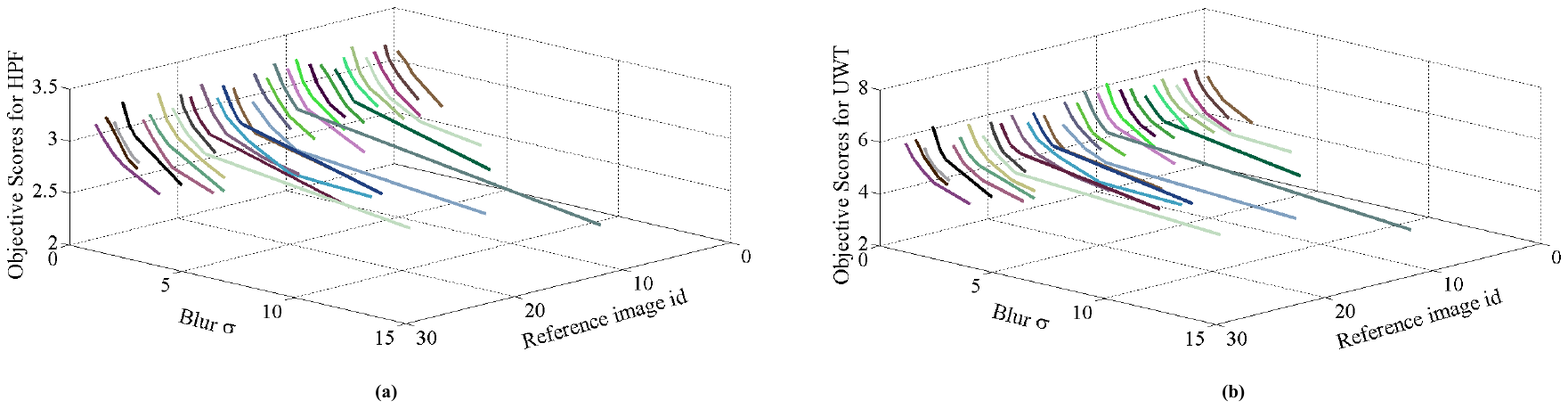}\\
  \caption{The variations of the objective scores with the Gaussian blur $\sigma$ in LIVE database using HPF and UWT.}\label{Fig:BlurSigmaLevel}\end{center}
\end{figure*}

In the second part of the experiment, we use images with varying levels of blur or lower sharpness generated from different pristine images. The LIVE database is chosen for this experiment. There are 29 reference or pristine quality images present in the LIVE database. Several images having varying Gaussian blur levels are generated from each of these reference images. The decrease of the value of the proposed measures with the increase of Gaussian blur is depicted in Fig.~\ref{Fig:BlurSigmaLevel}. Here, the values of the proposed measure are plotted against the standard deviation $\sigma$ of the Gaussian blur applied to the pristine image. As seen from the figure, from each of the 29 reference images, the blurred images are derived by gradually increasing $\sigma$ values. Thus, the subjective scores goes down gradually, indicating loss of sharpness and perceptual quality. The objective scores obtained from both HPF and UWT decreases gradually with  the increasing $\sigma$ values. Therefore, the objective scores are able to represent the loss of perceptual sharpness in each set of degraded images derived from the same pristine image. Though it is evident that perceived sharpness depends a lot on the image content~\cite{VuS32012,HassenSharpness2013}, for the high levels of blur, the scores are low signifying greater degradation. We also find that for HPF and UWT the ranges of sharpness score vary significantly.

\subsection{Performance in four databases}
\label{ssec:databasesBlurNRF}
\begin{figure*}
\begin{center}
  \includegraphics[width=\textwidth]{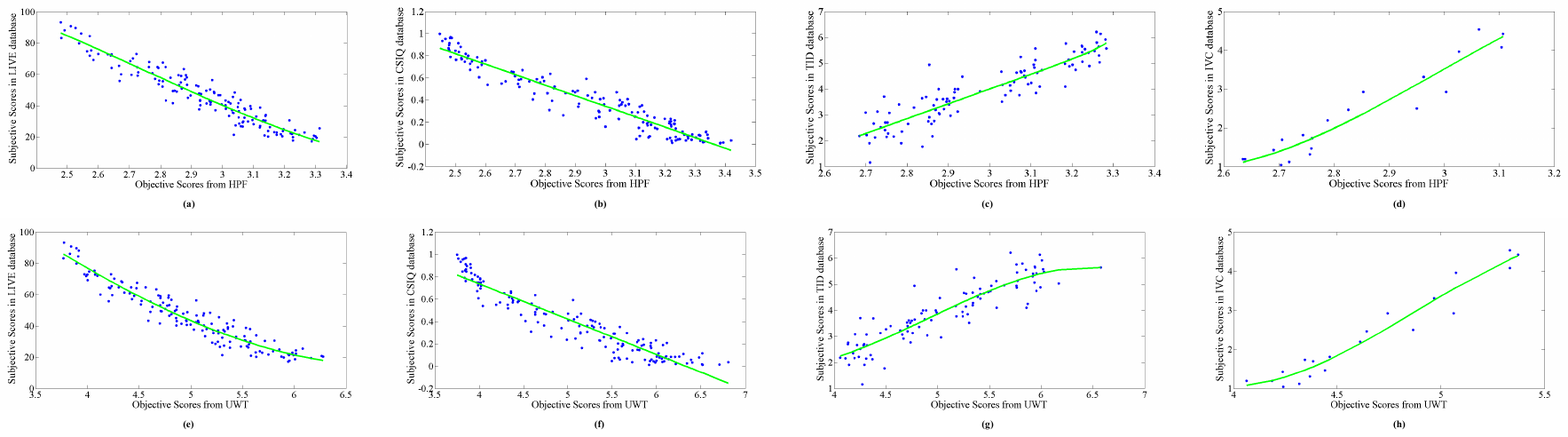}\\
  \caption{Scatter plots between the objective and subjective scores in blue. The green line represents the fitted values after applying the logistic function. }\label{Fig:Fitting_4_Databases}
\end{center}
\end{figure*}

In this experiment, the subjective scores in four databases are plotted against the objective scores obtained from the measures for different images as shown in Fig.~\ref{Fig:Fitting_4_Databases}. The scatter plot shows that a high degree of correlation exists between subjective values and the scores obtained by the proposed method. This experiment takes into account the performance across several images of varying content against the subjective scores. The scores mostly follow a monotonic variation with the subjective scores of all the blurred images taken together in a database for both HPF and UWT. A global quantitative evaluation is required to compare this performance with the existing methods and this is provided in the next section in comparison with the other methods.
\subsection{Comparisons with other methods}
\label{ssec:ComparisonResults}
We compare the proposed method in four databases against several FR and NR quality measures. Peak signal-to-noise ratio (PSNR) and  SSIM~\cite{WangSSIM04} are the two FR-IQA techniques we compare with. Considering the NR methods, we have divided them as NR-Specific and NR-General. In this context, NR-Specific means an NR quality measure that works only on blurriness or sharpness related distortion. The NR-specific methods used in comparison are : H-metric~\cite{ZhuBlur2009}, Q-metric~\cite{ZhuQmetric2010}, JNBM~\cite{FerzliBlur2009}, CPBD~\cite{NarvekarCPBD2011}, S3~\cite{VuS32012} and LPC-SI~\cite{HassenSharpness2013}. NR-General means the IQA technique is able to the assess the image quality for different distortions. BRISQUE~\cite{MittalBRISQUE2012} and BLIINDS~\cite{SaadBLIINDS2011} are the NR-General methods used for comparison here. The qualitative results obtained from the comparison process are presented in Table~\ref{Tab:Comparison}. The best performer's score in each database and for each measure is highlighted in bold and the best measure among the NR-specific methods is underlined.  For LIVE database, we find that SSIM has highest SROCC score, UWT has highest PLCC and lowest RMSE values and the proposed method provides highest KROCC value. In CSIQ databases, the proposed methods using HPF and UWT perform remarkably better than all the methods for all types of measures. For TID database, SSIM is the best performer for all measures but considering NR-specific methods only, the proposed method improves the performance over the existing ones to a large extent. For IVC database, UWT, LPC-SI and H-metric achieves the best performance using different measures.

We also note the average performances of the measures in all four databases presented in Tables~\ref{Tab:DAvgComparison} and~\ref{Tab:WAvgComparison}. A direct computation of the average values for SROCC, PLCC and KROCC reveals the proposed UWT based method has the best performance among all the measures. However, when the values are weighted by the number of relevant images present in the corresponding dataset, SSIM shows the best performance with SROCC. The proposed method based on UWT performs the best with PLCC while using KROCC, HPF is the best performer. The NR-general techniques work on a lot of distortions and hence achieving the best performance on a specific kind of distortion is difficult for them.
\begin{table*}
  \centering
  \caption{Comparison with the existing techniques}\label{Tab:Comparison}
  \setlength{\tabcolsep}{0.1cm}
  \begin{tabular}{ |c|c|cc|cccccccc|cc| }
  \hline
  &&\multicolumn{2}{c|}{FR}&\multicolumn{8}{c|}{NR-Specific}&\multicolumn{2}{c|}{NR-General}\\
  \hline
Databases&Measure&PSNR&SSIM\cite{WangSSIM04}&H-metric\cite{ZhuBlur2009}&Q-metric\cite{ZhuQmetric2010}&JNBM\cite{FerzliBlur2009}&CPBD\cite{NarvekarCPBD2011}&S3\cite{VuS32012}&LPC-SI\cite{HassenSharpness2013}&HPF&UWT&BRISQUE\cite{MittalBRISQUE2012}&BLIINDS II\cite{SaadBLIINDS2011}\\
\hline
\multirow{4}{*}{LIVE}&SROCC&0.7823&\textbf{0.9516}&0.6865&0.5482&0.7872&0.9182&0.9436&0.9389&\underline{0.9515}&0.9510&Used&Used\\
&PLCC&0.7748&0.9483&0.6977&0.6559&0.7835&0.9119&0.9525&0.9315&0.9537&\underline{\textbf{0.9559}}&for& for\\
&KROCC&0.5847&0.8011&0.5080&0.3971&0.6071&0.7632&0.7996&0.7785&\underline{\textbf{0.8094}}& 0.8086&training& training\\
&RMSE&11.6815&5.8634&13.2389&13.9424&11.5085&7.5852&5.6269&6.7352&5.5578&\underline{\textbf{5.4270}}&&\\
\hline
\multirow{4}{*}{CSIQ}&SROCC&0.9291&0.9245&0.7833&0.6528&0.7624&0.9182&0.9058&0.9068&\underline{\textbf{0.9495}}&0.9444&0.9032&0.8765\\
&PLCC&0.9081&0.8903&0.7467&0.6562&0.7887&0.9119&0.8921&0.9255&\underline{\textbf{0.9594}}&0.9449&0.9252&0.8930\\
&KROCC&0.7543&0.7657&0.6081&0.4860&0.5971&0.7632&0.7290&0.7197&\underline{\textbf{0.8053}}&0.7972&0.7353&0.6783\\
&RMSE&0.1201&0.1328&0.1910&0.2165&0.1780&7.5852&0.1296&0.1086&\underline{\textbf{0.0809}}&0.0939&0.1090&0.1290\\
\hline
\multirow{4}{*}{TID}&SROCC&0.8682&\textbf{0.9596}&0.5275&0.3276&0.6667&0.8412&0.8418&0.8561&\underline{0.9079}&0.9061&0.7989&0.8205\\
&PLCC&0.8666&\textbf{0.9555}&0.5553&0.2930&0.6535&0.8320&0.8540&0.8562&\underline{0.9043}&0.9014&0.7388&0.8260\\
&KROCC&0.7328&\textbf{0.8288}&0.3398&0.2210&0.4951&0.6310&0.6124&0.6362&\underline{0.7365}&0.7300&0.6229&0.6245\\
&RMSE&0.5953&\textbf{0.3465}&0.9766&1.1220&0.8921&0.6519&0.6110&0.6066&\underline{0.5011}&0.5081&0.8239&0.6614\\
\hline
\multirow{4}{*}{IVC}&SROCC&0.7893&0.8691&\underline{\textbf{0.9541}}&0.9323&0.6659&0.7690&0.8691&0.9398&0.9345&\underline{\textbf{0.9541}}&0.8239&0.5262\\
&PLCC&0.8527&0.9149&0.9627&0.9360&0.7037&0.8469&0.9279&\underline{\textbf{0.9736}}&0.9555&0.9728&0.8489&0.7806\\
&KROCC&0.6349&0.7090&0.8254&0.7831&0.4974&0.6138&0.7090&0.8042&0.8148&\underline{\textbf{0.8466}}&0.6561&0.3979\\
&RMSE&0.6309&0.4695&0.3093&0.4097&0.8138&0.6201&0.4265&\underline{\textbf{0.2607}}&0.3367&0.2646&0.6151&0.7136\\
\hline
\end{tabular}
\end{table*}

\begin{table*}
\centering
\caption{Direct averages over four databases}\label{Tab:DAvgComparison}
\setlength{\tabcolsep}{0.1cm}
\begin{tabular}{ |c|cc|cccccccc|cc| }
\hline
&\multicolumn{2}{c|}{FR}&\multicolumn{8}{c|}{NR-Specific}&\multicolumn{2}{c|}{NR-General}\\
\hline
Measure&PSNR&SSIM\cite{WangSSIM04}&H-metric\cite{ZhuBlur2009}&Q-metric\cite{ZhuQmetric2010}&JNBM\cite{FerzliBlur2009}&CPBD\cite{NarvekarCPBD2011}&S3\cite{VuS32012}&LPC-SI\cite{HassenSharpness2013}&HPF&UWT&BRISQUE\cite{MittalBRISQUE2012}&BLIINDS II\cite{SaadBLIINDS2011}\\
\hline
SROCC&0.8422&0.9262&0.7379&0.6152&0.7206&0.8617&0.8901&0.9104&0.9359&\underline{\textbf{0.9389}}&0.8420&0.7411\\
PLCC&0.8506&0.9273&0.7406&0.6353&0.7324&0.8757&0.9066&0.9217&0.9309&\underline{\textbf{0.9438}}&0.8376&0.8332\\
KROCC&0.6767&0.7762&0.5703&0.4718&0.5492&0.6928&0.7125&0.7347&0.7915&\underline{\textbf{0.7956}}&0.6714&0.5669\\
\hline
\end{tabular}
\end{table*}

\begin{table*}
\centering
\caption{Weighted averages over four databases}\label{Tab:WAvgComparison}
\setlength{\tabcolsep}{0.1cm}
\begin{tabular}{ |c|cc|cccccccc|cc| }
\hline
&\multicolumn{2}{c|}{FR}&\multicolumn{8}{c|}{NR-Specific}&\multicolumn{2}{c|}{NR-General}\\
\hline
Measure&PSNR&SSIM\cite{WangSSIM04}&H-metric\cite{ZhuBlur2009}&Q-metric\cite{ZhuQmetric2010}&JNBM\cite{FerzliBlur2009}&CPBD\cite{NarvekarCPBD2011}&S3\cite{VuS32012}&LPC-SI\cite{HassenSharpness2013}&HPF&UWT&BRISQUE\cite{MittalBRISQUE2012}&BLIINDS II\cite{SaadBLIINDS2011}\\
\hline
SROCC&0.8564&\textbf{0.9398}&0.6961&0.5514&0.7434&0.8925&0.9018&0.9074&\underline{0.9395}&0.9379&0.8587&0.8298\\
PLCC&0.8489&0.9275&0.6939&0.5821&0.7502&0.8895&0.9057&0.9132&0.9262&\underline{\textbf{0.9396}}&0.8505&0.8599\\
KROCC&0.6841&0.7905&0.5189&0.4054&0.5712&0.7241&0.7246&0.7242&\underline{\textbf{0.7906}}&0.7874&0.6878&0.6376\\

\hline
\end{tabular}
\end{table*}

\section{Conclusion and Future Works}
\label{sec:Conclusion}
We presented a technique for automatic assessment of perceived sharpness in natural images. The technique emphasizes on the application of high frequency content with proper exponent as the stimulus and combining it with the contrast. The logarithm of this combination is used to generate the sharpness map and sharpness score. The performance of the proposed method is demonstrated to provide improved performance over the state-of-the-art methods. The variation in the performance of the proposed technique is also analyzed for varying values of the exponent and for two types of high frequency content extraction methods using high pass filter and undecimated wavelet transform. We have divided the experiments into several distinct parts in order to properly visualize the consistency of performance. A maximum  pooling strategy has been adopted in our approach to arrive at the perceptual score. The future work lies in studying the effect of different pooling strategies for perceived sharpness evaluation.


\section*{Acknowledgments}
The work is supported in part by the Canada Research Chair program, and the Natural Sciences and Engineering Research Council of Canada.

\bibliographystyle{IEEEtran}
\bibliography{refs}

%

\end{document}